\documentclass[runningheads]{llncs}
\usepackage{graphicx}

\usepackage{tikz}
\usepackage{comment}
\usepackage{amsmath,amssymb} 
\usepackage{color}

\usepackage{hyperref}

\begin{document}
\pagestyle{headings}
\mainmatter
\def\ECCVSubNumber{7}

\title{Using Sentences as Semantic Representations in Large Scale Zero-Shot Learning}

\titlerunning{Using Sentences as Semantic Representations in Large Scale ZSL}
%
\author{Yannick Le Cacheux\inst{1,2} \and
Hervé Le Borgne\inst{1} \and
Michel Crucianu\inst{2}}
\authorrunning{Y. Le Cacheux et al.}
%
\institute{Université Paris-Saclay, CEA, List, F-91120, Palaiseau, France \\
\email{\{yannick.lecacheux,herve.le-borgne\}@cea.fr} \and
CEDRIC -- CNAM, Paris, France \\
\email{michel.crucianu@cnam.fr}}
\maketitle


\begin{abstract}
Zero-shot learning aims to recognize instances of unseen classes, for which no visual instance is available during training, by learning multimodal relations between samples from seen classes and corresponding class semantic representations.
These class representations usually consist of either attributes, which do not scale well to large datasets, or word embeddings, which lead to poorer performance. A good trade-off 
could be to employ short sentences in natural language as class descriptions. 
We explore different solutions to use such short descriptions in a ZSL setting and show that while simple methods cannot achieve very good results with sentences alone, a combination of usual word embeddings and sentences can significantly outperform current state-of-the-art
\footnote{A short version of this article was published in the Task-CV workshop @ECCV 2020}.
\end{abstract}


\section{Introduction and Related Work}

Zero-shot learning (ZSL) is useful when no visual samples are available for certain classes, provided we have semantic information for these classes~\cite{lampert2013attribute}.
It is then possible to train a model to learn the relations between the visual and semantic features using \emph{seen} classes, for which both modalities are available; these relations can later be employed to classify instances of \emph{unseen} classes, for which no visual sample is available during training, based on their semantic \emph{class prototypes}\cite{norouzi2013conse,frome2013devise,romeraparedes2015eszsl,lecacheux2018mmm,lecacheux2019iccv}. 
The semantic information can consist of vectors of attributes, 
e.g.\ binary codes for ``has fur'', ``has stripes'', etc.\ if classes are animal species. 
In a large-scale setting, it can be impractical to devise and provide attributes for hundreds or even thousands of classes. Word embeddings are then typically used to represent classes. However, a large performance gap still exists between these two types of class representations~\cite{lecacheux2020webly}.

An ideal solution could be to use short natural sentences to describe each class, as this is less time-consuming than providing comprehensive attributes and can be more visually informative than word embeddings derived from generic text corpora. 
The use of sentences as class descriptions in ZSL is not well studied.
Some works already employed sentences for ZSL, but not in a convenient setting. For instance, Akata \textit{et al.}~\cite{akata2015evaluation} rely on 10 sentences \emph{per image} instead of 1 per class.
Ba \textit{et al.}~\cite{lei2015predicting} and Elhoseiny \textit{et al.}~\cite{elhoseiny2017link} use tf-idf from large text collections such as Wikipedia articles and different neural architectures to obtain semantic representations. Zhu \textit{et al.}~\cite{zhu2018generative} generate visual features from noisy text description (again from Wikipedia articles) using Generative Adversarial Network.
These works as well as others such as \cite{elhoseiny2013write,qiao2016less} tend to focus on fine-grained recognition and explicitly or implicitly assume that classes are similar (consisting for example of birds species) and that class descriptions will contain specific information (for example regarding the beak, the wings, the plumage...). On the other hand, we focus on generic object recognition and make close to no assumption regarding the nature of these descriptions.
The closest work to ours is probably Hascoet \textit{et al.}~\cite{hascoet2019semantic}, in which different methods to obtain prototypes from WordNet definitions are evaluated, but reported performance is significantly below that of usual word embeddings.

In this article, we explore several ideas to leverage sentences to build semantic embeddings for ZSL and show that this can significantly outperform previous best reported performance. These proposals are easy to implement and computationally light. We provide the code\footnote{\url{https://github.com/yannick-lc/zsl-sentences}} as well as the corresponding embeddings which can be used out-of-the-box as better quality semantic representations.


\section{Proposed Method}
To deal with a zero-shot learning (ZSL) task, one considers a set $\mathcal{C}_s$ of \emph{seen} classes used during training and a set $\mathcal{C}_u$ of \emph{unseen} classes that are available for the test only, with $\mathcal{C}_s \cap \mathcal{C}_u = \emptyset$. Each class is also associated to a semantic \emph{class prototype} $\mathbf{s}_c\in\mathbb{R}^K$ that characterizes it. 
Let us consider a training set $\{(\mathbf{x}_i,y_i), i=1\dots N\}$ with labels $y_i\in\mathcal{C}_s$ and visual features $\mathbf{x}_i\in\mathbb{R}^D$. The goal of the ZSL task is to learn a compatibility function $f: \mathbb{R}^D \times \mathbb{R}^K \rightarrow \mathbb{R}$ assigning a similarity score to a visual sample $\mathbf{x}$ and a class prototype $\mathbf{s}$. It is usually a parametrized function that is learned by minimizing a regularized loss $\mathcal{L}$:
\begin{equation} \label{eq:objective}
\frac{1}{N} \sum_{i=1}^{N} \sum_{c=1}^{|\mathcal{C}_s|} \mathcal{L}(f(\mathbf{x}_i, \mathbf{s}_c), y_i) + \lambda\Omega[f]
\end{equation}
where $\Omega$ is a regularization term weighted by $\lambda$. Once $f$ is learned, the testing phase consists in determining the label $\hat{y}\in\mathcal{C}_u$  corresponding to a visual sample $\mathbf{x}$ such that 
$\hat{y} = \underset{c \in \mathcal{C}_u}{\text{arg\,max }} f(\mathbf{x}, \mathbf{s}_c)$.

A simple linear model is presented in Section~\ref{sec:linear_model}. Our contribution presented in Section~\ref{sec:contribution} consists in leveraging sentences to build semantic prototypes $\mathbf{s}_c$

\subsection{Linear ZSL model.}\label{sec:linear_model}
We adopt a linear ridge regression model, in which semantic prototypes are projected into the visual features space so as to minimize a regularized least square loss. This choice is motivated by the good and consistent results of this model in \cite{lecacheux2020webly}, as well as its low computational cost and ease of reproducibility. The choice of the visual space as the projection space is based on \cite{shigeto2015hubness}, which argues that this choice enables to mitigate the hubness problem \cite{radovanovic2010hubs}.
Results with other models from \cite{hascoet2019zero} are provided in Section~\ref{sec:results_supp}.

The visual features $\{\mathbf{x}_1, \dots, \mathbf{x}_N\}$ are extracted with a pre-trained network, so that $\mathbf{x}_n \in \mathbb{R}^{D}$, $D$ the dimension of the visual space. We thus have a corresponding training matrix $\mathbf{X} \in \mathbb{R^{N \times D}}$.
The semantic representations $\{\mathbf{s}_1, \dots, \mathbf{s}_C\}$ of the $C$ seen (training) classes lie in $\mathbb{R}^K$.
We write $\mathbf{T}=(\mathbf{t}_1, \dots, \mathbf{t}_N)^\top \in \mathbb{R}^{N \times K}$ the semantic representations associated with each training sample $\mathbf{x}_n$ with label $y_n$ so that $\mathbf{t}_n=\mathbf{s}_{y_n}$.

The resulting loss for the ridge regression with parameters $\boldsymbol{\Theta} \in \mathbb{R}^{K \times D}$ is
\begin{equation} \label{eq:ridge_loss}
\frac{1}{N} \lVert \mathbf{X - \mathbf{T}\boldsymbol{\Theta}} \rVert_F^2 + \lambda \lVert \boldsymbol{\Theta} \rVert_F^2
\end{equation}
where $\lambda$ a hyperparameter weighing the $\ell2$ regularization penalty $\lVert \boldsymbol{\Theta} \rVert_F^2$, and $\lVert \cdot \rVert_F$ is the Frobenius norm.
This minimization problem has a closed-form solution:
\begin{equation} \label{eq:ridge_solution}
\boldsymbol{\Theta} = (\mathbf{T}^\top\mathbf{T} + \lambda N \mathbf{I}_K)^{-1} \mathbf{T}^\top\mathbf{X}
\end{equation}

During the prediction phase, given a testing sample $\mathbf{x}$ and the semantic representations $\{\mathbf{s'}_1, \dots, \mathbf{s'}_{C'}\}$ of the $C'$ unseen classes, $\mathbf{s'}_{c'} \in \mathbb{R}^K$, we predict label $\hat{y}$ corresponding to the class whose prototype's projection is closest to $\mathbf{x}$:
\begin{equation}
\hat{y} = \underset{c'}{\text{argmin }} \lVert \mathbf{x} - \boldsymbol{\Theta}^\top \mathbf{s}_{c'} \rVert_2
\end{equation}

\subsection{Sentence-Based Embeddings}\label{sec:contribution}
A simple baseline consists in averaging the embeddings of the words in the definition, as is usually done for class names consisting of several words:
if a sentence $s$ describing a class has $N$ words with respective embeddings $\{\mathbf{w}_1, \dots, \mathbf{w}_N\}$, then the corresponding semantic representation is $\mathbf{s} = \frac{1}{N} \sum_{n=1}^N \mathbf{w}_n$.
We call this baseline the \textbf{\textit{Def\textsubscript{average}}} approach.
However, as illustrated in Fig.~\ref{fig:attention}, not all words are equally important in a short sentence description. We therefore explore the use of attention mechanisms: the sentence embedding is a weighted average of the embeddings of its words, so that more important words contribute more to the result.
We consider two ways to achieve this.

\subsubsection{Visualness Based Approach}
The first, called the \textbf{\textit{Def\textsubscript{visualness}}} approach, aims to 
estimate how ``visual'' a word is. For a given word $w_n$, we thus collect the 100 
most relevant images from Flickr using the website's search ranking.
We obtain representations $\{\mathbf{r}_1^i, \dots, \mathbf{r}_{M}^i\}$ with 2048 dimensions ($\mathbf{r}_m\in \mathbb{R}^{2048}$) using a pre-trained ResNet (see Section~\ref{sec:features}).
We then measure the average distance of vectors $\mathbf{r}_m^i$ to the mean vector $\bar{\mathbf{r}}^i = \frac{1}{M} \sum_{m=1}^M \mathbf{r}_m^i$ to obtain the additive inverse of the visualness $v^i$:
\begin{equation} \label{eq:visualness}
v^i = - \frac{1}{M} \sum_{m=1}^M \lVert \mathbf{r}_m^i - \bar{\mathbf{r}^i} \rVert_2
\end{equation}
We hypothesize that words with high visual content have features close to each other, so that the quantity $v^i$ from Equation~\ref{eq:visualness} can be used as a qualitative measure of the intuitive concept of how ``visual'' a given word is. Examples of words with high and low visualness shown in Section~\ref{sec:visualness} tend to confirm that this hypothesis is reasonable.

\begin{figure}
    \centering
    \includegraphics[width=250pt]{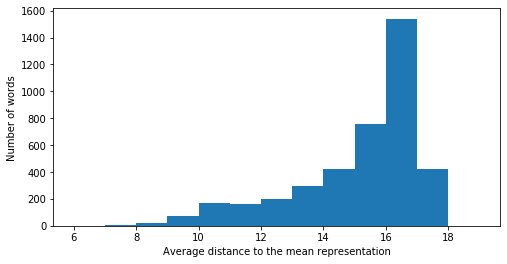}
    \caption{Inverse of the visualness (low values correspond to high visualness) for the 4059 words from class names and WordNet definitions.}
    \label{fig:hist_visualness}
\end{figure}

Figure~\ref{fig:hist_visualness} shows the distribution of the inverse of the visualness for the 4059 words from class names and WordNet definitions of classes from \cite{hascoet2019zero}. Equation~\ref{eq:visualness} is then used to obtain weights for each word based on its visualness in order to use a weighted average of the word embeddings from a definition as class prototype. As the initial scale of the average distances / negative visualnesses is arbitrary, a temperature $\tau$ is introduced in the softmax to weigh each word, so that the resulting sentence embedding is
\begin{equation} \label{eq:softmax}
\mathbf{s} = \sum_{n=1}^N \frac{\text{exp}(v_n / \tau)}{\sum_{k=1}^N \text{exp}(v_k / \tau)} \mathbf{w}_n
\end{equation}
The value $\tau$ is cross-validated on the validation set -- see Section~\ref{sec:implementation_details}. In practice, this often leads to selecting $\tau=5$.

\subsubsection{Learned Attention Based Approach}
A second approach called \textbf{\textit{Def\textsubscript{attention}}} aims to learn to predict the visualness $v^i$ of word $w^i$ from its embedding $\mathbf{w}^i \in \mathbb{R}^K$ such that $v^i = \boldsymbol{\theta}^\top \mathbf{w}^i$, where $\boldsymbol{\theta}$ are learned parameters.
Equation~\ref{eq:visualness} can then be used to create a prototype from a class' definition.
As the $v^i$ are directly learned, it is no longer necessary to account for their initial scale. We can thus discard the temperature in the softmax by setting $\tau=1$ in Equation~\ref{eq:visualness}.

Different ways could be considered to learn $\boldsymbol{\theta}$. A straightforward approach could consist in randomly initializing $\boldsymbol{\theta}$ along with $\boldsymbol{\Theta}$ from Equation~\ref{eq:ridge_loss}, computing visualnesses $v^i = \boldsymbol{\theta}^\top \mathbf{w}^i$, computing class prototypes $\mathbf{s}_c$ using the $v^i$ and Equation~\ref{eq:visualness}, computing $\mathbf{T}$ and finally computing the loss from Equation~\ref{eq:ridge_loss}; we could then use back-propagation and gradient descent to update $\boldsymbol{\theta}$ and $\boldsymbol{\Theta}$ until convergence.

However, we instead take advantage of the closed-form solution from Equation~\ref{eq:ridge_solution} and proceed as follows: we randomly initialize $\boldsymbol{\theta}$ and compute class prototypes $\mathbf{s}_c$ and $\mathbf{T}$ using Equation~\ref{eq:visualness} as previously. We then directly estimate $\boldsymbol{\Theta}$ using the closed-form solution from Equation~\ref{eq:ridge_solution}, and use this value to compute the loss in Equation~\ref{eq:ridge_loss}. We then back-propagate the gradient and perform gradient descent on $\boldsymbol{\theta}$ only, the value of $\boldsymbol{\Theta}$ being estimated with Equation~\ref{eq:ridge_solution} at each iteration.
We repeat this process for 50 ``epochs''.
An illustration of the resulting weights from both approaches is shown in Fig.~\ref{fig:attention}.

The resulting sentence embeddings can then be compared to the standard class prototypes obtained by embedding the 
class name (\textbf{\textit{Classname}} approach).
Since recent results show that hierarchical and graph relations between classes contain valuable information \cite{hascoet2019zero}, in the \textbf{\textit{Classname+Parent}} approach we combine the \textit{Classname} prototype with the prototype of its parent class.

Finally, we experiment with different combinations of the base approaches:
\begin{itemize}
    \item \textbf{\textit{Classname+Def\textsubscript{average}}},
    \item \textbf{\textit{Classname+Def\textsubscript{visualness}}}
    \item \textbf{\textit{Classname+ Def\textsubscript{visualness} +Parent}}.
\end{itemize}
All these combinations simply consist in finding a value $\mu \in [0,1]$ such that the combined prototype is $\mathbf{s} = \mu \mathbf{s}_1 + (1-\mu) \mathbf{s}_2$, given $\mathbf{s}_1$ and $\mathbf{s}_2$ the prototypes of two approaches.


\section{Experiments}
\subsection{Settings}\label{sec:implementation_details}
\subsubsection{Selecting hyperparameters.}
Hyperparameter values such as $\lambda$ in Equation~\ref{eq:ridge_loss} or $\tau$ in Equation~\ref{eq:visualness} are selected by cross-validation on \emph{seen} classes, keeping 200 of the 1000 seen training classes as unseen validation classes. The model is then retrained on all 1000 training classes with the selected values.

When several prototypes $\mathbf{s}_1$ and $\mathbf{s}_2$ are combined with $\mathbf{s} = (1-\mu) \mathbf{s}_1 + \mu \mathbf{s}_2$,  $\mu \in [0,1]$, 
for instance in the \textit{Classname+Def\textsubscript{average}} or \textit{Classname+Def\textsubscript{visualness}} approaches, $\mu$ is also considered to be a hyperparameter and its value is selected by cross-validation along with the other hyperparameters. In these cases, it leads to surprising results with a frequently selected value of $\mu=0.7$, meaning the combined prototype is 70\% from the definition prototype and 30\% from the classname prototype, even though the definition alone performs far worse than the classname alone. Nonetheless, this still led to consistent results.

On the other hand, when combining a prototype with the prototype of its parent class, cross-validation tended to yield very inconsistent and unstable values.
When combining a prototype with its parent prototype, we therefore fixed $\mu=0.25$ (meaning 75\% base class and 25\% parent class) for all methods and all word embeddings.
Note that in the \textbf{\textit{Classname+Def\textsubscript{visualness}}} and \textbf{\textit{Classname+ Def\textsubscript{visualness} +Parent}} approach, the prototype from the parent class is also built with \textit{Classname+Def\textsubscript{visualness}}, and the value of $\mu$ is the same as the one selected for the child class.

\subsubsection{Pre-trained word embeddings.}
We used widely adopted pre-trained embeddings available on the Internet for Word2vec~\cite{mikolov2013w2v}, FastText~\cite{bojanowski-etal-2017-enriching} and Glove~\cite{pennington-etal-2014-glove}. We used a Word2vec embedding model pre-trained on Wikipedia as we found it gave better results than other version, such as a Word2vec embedding model trained on Google News. For the same reason, we used FastText and Glove models pre-trained on Common Crawl.
We used a 300-dimension version for all three. 
The pre-trained embedding models were downloaded from the following links:
\begin{itemize}
    \item Word2vec: \url{https://wikipedia2vec.github.io/wikipedia2vec/pretrained/} (English version trained on Wikipedia).
    \item Fasttext: \url{https://fasttext.cc/docs/en/english-vectors.html} (version trained on Common Crawl with 600B tokens, no subword information).
    \item Glove: \url{https://nlp.stanford.edu/projects/glove/} (version trained on Common Crawl with 840B tokens).
\end{itemize}

For Elmo~\cite{peters2018deep}, we similarly adopted a pretrained version obtained from \url{https://allennlp.org/elmo}. We used the original version with 93.6 million parameters, pre-trained on the 1 Billion Word Benchmark. Elmo embeddings have dimension $3 \times 1024$ (a 1024-dimensional embedding from each of the three layers). For the sake of simplicity we combine the three layers using the same weight of $0.33$ to obtain a single 1024-dimensional representation for each word, as weights fine-tuned for our specific task gave similar results.

\subsubsection{Visual and semantic features.} \label{sec:features}
Visual features are obtained from the last pooling layer of a ResNet~\cite{he2016deep} model pre-trained on ImageNet~\cite{deng2009imagenet}. They therefore have 2048 dimensions. We use the ResNet-101 implementation of PyTorch~\cite{paszke2019pytorch}. Visual features are normalized to have unit $\ell$2-norm.

\textit{Classname} ZSL embeddings are obtained from synsets as follows: a synset in WordNet~\cite{miller1995wordnet} can consist of several lemmas, each lemma consisting of several words. The prototype of a lemma is the mean of the embeddings of its words when such embeddings exist in the pre-trained model. The prototype of the synset and thus of the class is the mean of the prototypes of its lemmas.
In the approximately 6 (depending on the pre-trained embedding) cases where no word from any lemma had an embedding in the pre-trained model, we used the embedding from the word \textit{``thing''} as class prototype.
All semantic prototypes are $\ell$2-normalized, for the \textit{Classname} approach as well as all the others.

\subsubsection{Evaluation  protocol}
We measure ZSL accuracy on the large-scale ImageNet dataset, with synsets as classnames and corresponding WordNet definitions as sentences. We also use the WordNet hierarchy to determine parent classes. We employ the experimental protocol of~\cite{hascoet2019zero}, with the same ResNet features and, especially, the same train/test splits since~\cite{hascoet2019zero} evidenced significant structural bias with previous popular splits. 
Glove embeddings are employed in~\cite{hascoet2019zero}; we also compare with Word2vec and FastText \cite{almeida2019word}, as well as Elmo contextual embeddings using pre-trained embeddings available on the Internet (details are provided in the supplementary material). 
%
The ZSL model is a ridge regression from the semantic to the visual space as it gives good, consistent and easily reproducible results (see~\cite{lecacheux2020webly}). 

\begin{table}[t]
\begin{center}
\caption{Comparison of approaches on ImageNet with a linear model. Best previous reported result is 14.1 in~\cite{hascoet2019zero}; the result marked with * corresponds to the same setting (use of \textit{Classname} with Glove embeddings).} 
\label{tab:results}
\begin{tabular}{|l|ccc|c|}
\hline
& Word2vec & FastText & Glove & Elmo \\
\hline
\textit{Classname} & 12.4 & 14.8 & 14.5* & 10.9 \\
\textit{Classname+Parent} & 13.4 & 15.9 & 15.4 & 11.4 \\
\hline
\textit{Def\textsubscript{average}} & 9.7 & 10.6 & 10.0 & 8.7 \\
\textit{Def\textsubscript{visualness}} & 10.5 & 10.9 & 10.5 & 9.5 \\
\textit{Def\textsubscript{attention}} & 10.5 & 11.0 & 10.2 & 9.5 \\
\hline
\textit{Classname+Def\textsubscript{average}} & 14.6 & 17.2 & 16.9 & 12.2 \\
\textit{Classname+Def\textsubscript{visualness}} & 14.8 & 17.3 & 16.8 & 12.1 \\
\textit{Classname+Def\textsubscript{visualness}+Parent} & \textbf{15.4} & \textbf{17.8} & \textbf{17.3} & \textbf{12.5} \\
\hline
\end{tabular}
\end{center}
\end{table} 

\subsection{Results with Linear Model}
\begin{figure}
    \centering
    \includegraphics[width=\textwidth]{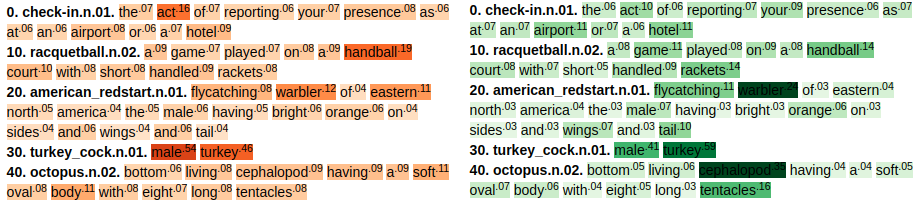}
    \caption{Illustration of attention scores on some test classes. \emph{Left}: weights from \textit{Def\textsubscript{visualness}} after softmax (the temperature is $\tau=5$ so differences are less pronounced than initially). \emph{Right}: weights learned with \textit{Def\textsubscript{attention}}, with FastText embeddings.}
    \label{fig:attention}
\end{figure}

Results are shown in Table~\ref{tab:results}.
The baseline \textit{Def\textsubscript{average}} approach performs poorly compared to the usual \textit{Classname} approach.
Attention provides a slight improvement (both \textit{Def\textsubscript{visualness}} and \textit{Def\textsubscript{attention}} give comparable results), but still does not match \textit{Classname}, even though attention weights seem reasonable (some examples are shown in Fig.~\ref{fig:attention}).
The use of parent information improves results compared with \textit{Classname} alone, which is consistent with~\cite{hascoet2019zero} where the best methods make use of hierarchical relations between classes.

The combination of \textit{Classname} and \textit{Def} approaches brings significantly better scores
than either separately.
Surprisingly, while any \textit{Def} alone has lower performance than \textit{Classname} alone, the best trade-off obtained by cross-validation is 70\% definition and 30\% classname in every case, meaning that the definition has a much stronger presence than the class name in the resulting embedding. 

The word embedding method has an impact on performance; FastText consistently ranks above the other methods. 
Elmo has surprisingly low performance including with fine-tuned attention, even though one could expect attention to be more effective here as Elmo considers the role of the word in the sentence. 

\subsection{Results with Other Models} \label{sec:results_supp}

We report results with the best performing prototypes for a few other ZSL models in addition to the linear model already described. We specifically report results for models used in \cite{hascoet2019zero} (except for graph-convolutional models which use a different setting).
For ConSE~\cite{norouzi2013conse}, DeViSE~\cite{frome2013devise} and ESZSL~\cite{romeraparedes2015eszsl}, we use the implementation from \cite{hascoet2019zero}. We report results averaged over 5 runs with different random initializations of parameters for ConSE and DeViSE. Since the other models use closed-form solutions, it is not necessary to report results averaged over several runs as there is no variance in the results.
We also provide results with another simple linear model, consisting in a ridge regression from the visual to the semantic space. This model is called Linear\textsubscript{$V \rightarrow S$}, as opposed to Linear\textsubscript{$S \rightarrow V$} which is the ridge regression model from the semantic to the visual space  described in Section~\ref{sec:linear_model}.

We report results with the best class prototypes as determined in Table~1 of the main paper, i.e. the ones obtained with the \textit{Classname+Def\textsubscript{visualness}+Parent} approach with the FastText embeddings. These prototypes will be made available to the community along with the others.
Top-1, top-5 and top-10 accuracies for the models mentioned above with these prototypes are provided in Table~\ref{tab:results_models}.

\begin{table}
\begin{center}
\caption{Top-k ZSL accuracy for different models, using the \textit{Classname+Def\textsubscript{visualness}+Parent} prototypes constructed from FastText embeddings.}
\label{tab:results_models}
\begin{tabular}{|l|ccc|}
\hline
& ~~top-1~~ & ~~top-5~~ & ~~top-10~~ \\
\hline
\textbf{Linear\textsubscript{$S \rightarrow V$}} & 17.8 & 43.6 & 56.7 \\
Linear\textsubscript{$V \rightarrow S$} & 9.1 & 26.2 & 36.7 \\
ESZSL & 16.3 & 40.6 & 52.4 \\
ConSE & 12.7 & 31.8 & 42.4 \\
DeViSE & 14.0 & 38.3 & 52.1 \\
\hline
\end{tabular}
\end{center}
\end{table}

\subsection{Illustration of visualness} \label{sec:visualness}

\begin{figure}[tb]
    \centering
    \includegraphics[width=350pt]{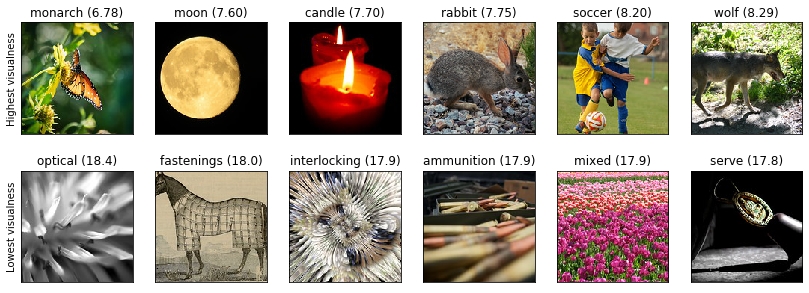}
    \caption{\emph{Top}: words with highest visualness. \emph{Bottom}: words with lowest visualness. The visualness of a word is the inverse of the mean distance (shown in parenthesis) to the mean representation of visual features from the top 100 corresponding images from Flickr, see Section~\ref{sec:visualness}. Top 1 image with no copyright restriction is displayed.}
    \label{fig:illustration_visualness}
\end{figure}

We provide a few examples of the words with the highest and lowest visualness in Figure~\ref{fig:illustration_visualness}, as well as the corresponding inverse visualness (the mean distance to the mean feature representations for images associated with this word) and the corresponding top image result with no copyright restriction from Flickr.

\section{Conclusion.}
To scale zero-shot learning to very large datasets it is important to solve the problem of providing class prototypes for many classes. The use of class name embeddings scales better than the provision manually-defined attributes but the resulting performance is nowhere near. We suggest that low-cost textual content, consisting in one sentence per class, can bring substantial performance improvements when combined with class name embeddings, when they are processed with the proposed approaches. 
The improved class prototypes for ImageNet are available at \url{https://github.com/yannick-lc/zsl-sentences}.


Beyond the case of zero-shot learning, other tasks are known to suffer from an incomplete semantic information, such as image retrieval, classification~\cite{znaidia2012icmr} or annotation~\cite{znaidia2013icmr}.  The work we presented in this article can thus benefit to these tasks, in particular when one addresses large-scale datasets. Predicting automatically the important words into sentences with regard to their visualness leads to semantic representation that can be mixed to visual representation to build intrinsic multimedia representations~\cite{znaidia2012icpr} that can be used to address cross-modal retrieval~\cite{tran16cvpr} and classification~\cite{tran16ivl}.

\bibliographystyle{splncs04}
\bibliography{egbib}
\end{document}